\documentclass[10pt,twocolumn,letterpaper]{article}

\usepackage{cvpr}
\usepackage{times}
\usepackage{epsfig}
\usepackage{graphicx}
\usepackage{amsmath}
\usepackage{amssymb}
\usepackage{epstopdf}
\usepackage{subfig}
\usepackage{multirow}

\usepackage[breaklinks=true,bookmarks=false]{hyperref}

\cvprfinalcopy 

\def\cvprPaperID{9255} 

\ifcvprfinal\fi
\begin{document}

\hyphenpenalty=5000
\tolerance=2000
\title{A Novel Recurrent Encoder-Decoder Structure for Large-Scale Multi-view Stereo Reconstruction from An Open Aerial Dataset}

\author{Jin Liu and Shunping Ji\thanks{Corresponding author} \\
School of Remote Sensing and Information Engineering, Wuhan University\\
{\tt\small \{liujinwhu, jishunping\}@whu.edu.cn}
}

\maketitle
\thispagestyle{empty}

\begin{abstract}
   A great deal of research has demonstrated recently that multi-view stereo (MVS) matching can be solved with deep learning methods. However, these efforts were focused on close-range objects and only a very few of the deep learning-based methods were specifically designed for large-scale 3D urban reconstruction due to the lack of multi-view aerial image benchmarks. In this paper, we present a synthetic aerial dataset, called the WHU dataset, we created for MVS tasks, which, to our knowledge, is the first large-scale multi-view aerial dataset. It was generated from a highly accurate 3D digital surface model produced from thousands of real aerial images with precise camera parameters. We also introduce in this paper a novel network, called RED-Net, for wide-range depth inference, which we developed from a recurrent encoder-decoder structure to regularize cost maps across depths and a 2D fully convolutional network as framework. RED-Net's low memory requirements and high performance make it suitable for large-scale and highly accurate 3D Earth surface reconstruction. Our experiments confirmed that not only did our method exceed the current state-of-the-art MVS methods by more than 50\% mean absolute error (MAE) with less memory and computational cost, but its efficiency as well. It outperformed one of the best commercial software programs based on conventional methods, improving their efficiency 16 times over. Moreover, we proved that our RED-Net model pre-trained on the synthetic WHU dataset can be efficiently transferred to very different multi-view aerial image datasets without any fine-tuning. Dataset and code are available at \url{ http://gpcv.whu.edu.cn/data}.
\end{abstract}

\section{Introduction}
Large-scale and highly accurate 3D reconstruction of the Earth's surface, including cities, is mainly realized from dense matching of multi-view aerial images implemented and dominated by commercial software such as Pix4D~\cite{RN01}, Smart3D~\cite{RN02}, and SURE~\cite{RN03}, all of which were developed from conventional methods~\cite{RN04,RN05,RN06}. Recent attempts at multi-view stereo (MVS) matching with deep learning methods are found in the literature~\cite{RN07,RN08,RN09,RN10,RN11}. While these deep learning approaches can produce satisfactory results on close-range object reconstruction, they have two critical limitations when applied to Earth surface reconstruction from multi-view aerial images. The first limitation is the lack of aerial dataset benchmarks, which makes it difficult to train, discover, and improve the appropriate networks through between-method comparison. In addition, most of the existing MVS datasets are images of laboratory, and models trained on them cannot be satisfactorily transferred to a bird's eye view of a terrestrial scene. The second limitation of these methods is their high GPU memory demand in recent MVS networks~\cite{RN09,RN11,RN12,RN13}, which makes them less suitable for large-scale and high-resolution scene reconstruction. The state-of-the-art R-MVSNet method~\cite{RN10} has achieved depth inference with unlimited depth-wise resolution, however, the resolution quality of its results is not high as the output depth map is down-sampled four times.\\
In this paper, we present a synthetic aerial dataset we created for large-scale MVS matching and Earth surface reconstruction. Each image in the dataset was simulated from a complete and accurate 3D urban scene produced from a real multi-view aerial image collection with software and careful manual editing. The dataset includes thousands of simulated images covering an area of $6.7\times2.2$ km$^2$, along with the ground truth depth and camera parameters for multi-view images, as well as disparity maps for rectified epipolar images. Due to the large size of the aerial images ($5376\times5376$ pixels), there are subsets provided consisting of cropped sub-blocks that can be used directly for training CNN models on a single GPU. Note that the simulated camera parameters are unbiased and the provided ground truths are absolutely complete even in occluded regions, which ensures the accuracy and reliability of the dataset for detailed 3D reconstruction.\\
We also introduce in this paper an MVS network, called RED-Net, we created for large scale MVS matching. A recurrent encoder-decoder (RED) architecture is utilized to sequentially regularize cost maps obtained from a series of convolutions on multi-view images. When compared to the state-of-the-art method~\cite{RN10}, we achieved higher efficiency and accuracy using less GPU memory while maintaining unlimited depth resolution, which is beneficial to city-scale reconstruction. Our experiments confirmed that RED-Net outperformed all the comparable methods evaluated on the WHU aerial dataset.\\
We had a third aim for our work beyond addressing the two limitations of the existing methods. That goal was to demonstrate that our MVS network could be generalized for cross-dataset transfer learning. We demonstrate here that RED-Net pre-trained on our WHU dataset could be directly applied on another quite different aerial dataset with slightly better accuracy than one of the best commercial software programs with efficiency improved 16 times over.
\section{Related Work}
\subsection{Datasets}
{\bf Two-view datasets}. Middlebury~\cite{RN14} and KITTI~\cite{RN15} are two popular datasets for stereo disparity estimation. However, these datasets are too small for current applications, especially when training deep learning models, and the lack of sufficient samples often leads to overfitting and low generalization. Considering this situation,~\cite{RN16} created a large synthetic dataset that consists of three subsets: FlyingThings3D, Monkaa, and Driving, which provide thousands of stereo images with dense and complete ground truth disparities. However, a model pre-trained on this synthetic dataset cannot easily be applied to a real scene dataset due to the heterogeneous data sources.\\
\indent {\bf Multi-view datasets}. The Middlebury multi-view dataset~\cite{RN17} was designed for evaluating MVS matching algorithms on equal ground and is a collection of calibrated image sets from only two small scenes in a laboratory environment. The DTU dataset~\cite{RN18} is a large scale close-range MVS benchmark that contains 124 scenes with a variety of objects and materials under different lighting conditions, which make it well-suited for evaluating advanced methods. The Tanks and Temples benchmark~\cite{RN19} provides high-resolution data with large-size images acquired in complex outdoor environments. A recent benchmark called ETH3D~\cite{RN20} was created for high-resolution stereo and multi-view reconstruction, which consists of artificial scenes and outdoor and indoor scenes and represents various real-world reconstruction challenges.\\
Reconstructing the Earth's surface and cities is mainly realized with matching multi-view aerial images. The ISPRS Association and the EuroSDR Center jointly provided two small aerial datasets called M\"unchen and Vaihingen~\cite{RN21}, which consist of dozens of aerial images; however, these datasets are currently not publicly accessible. In our work, we created a large-scale synthetic aerial dataset with accurate camera parameters and complete ground truths for MVS method evaluation and urban scene reconstruction.
\subsection{Networks}
\begin{figure*}[htp]
\begin{center}
\includegraphics[width=0.75\linewidth]{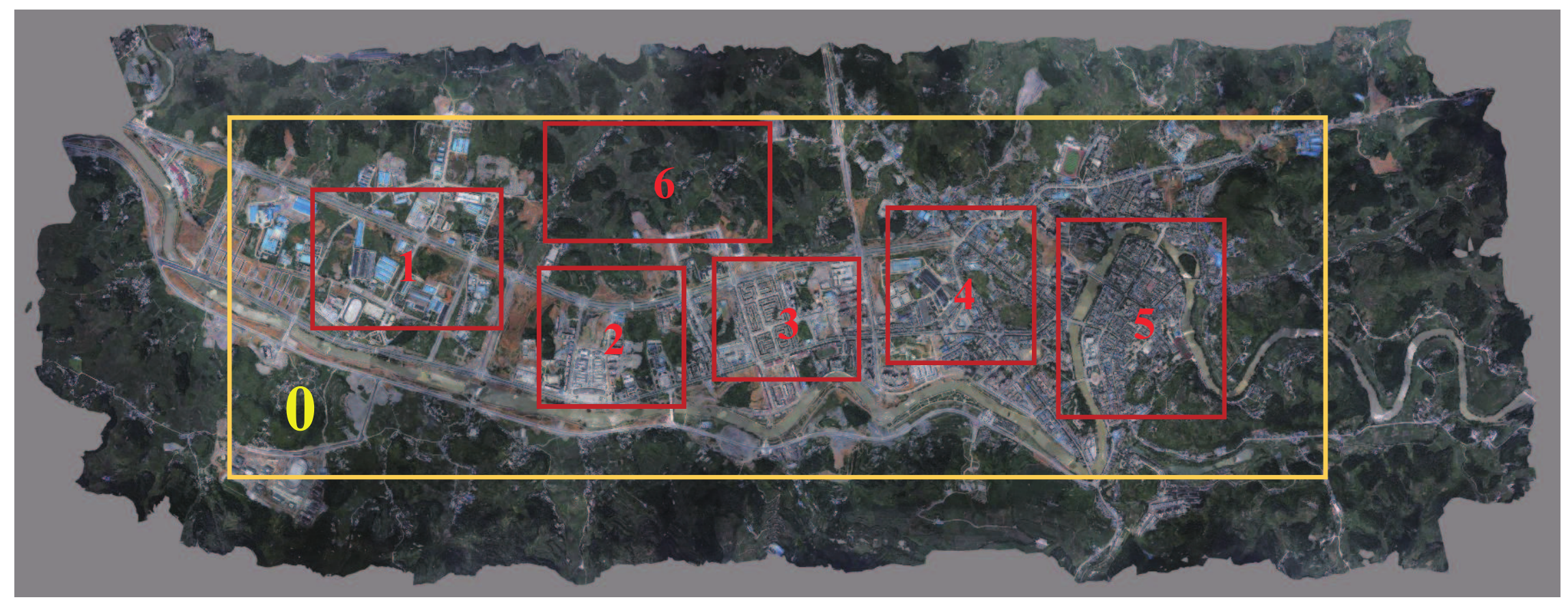}
\end{center}
\setlength{\abovecaptionskip}{-0.3cm}
   \caption{The dataset. Area 0: the complete dataset consists of 1,776 virtual aerial images each $5376\times5376$ pixels in size. For facilitating machine learning methods, areas 1/4/5/6 were allocated for the training set, which consisted of 261 images. Areas 2 and 3, which consisted of 93 images, were used as the test set. In the training and testing area, the images also were cropped into tiles of $768\times384$ pixel-size for a single GPU.}
\label{fig:1}
\vspace{-0.3cm}
\end{figure*}
Inspired by the success of the deep learning based stereo methods~\cite{RN22,RN23,RN24,RN25}, some researchers attempted to apply CNNs to the MVS task. Hartmann et al.~\cite{RN26} proposed an N-way Siamese network to learn the similarity score over a set of multi-patches. The first end-to-end learning network designed for MVS was SurfaceNet~\cite{RN11} by building colored voxel cubes outside the network to encode the camera parameters through perspective projection, which combined multi-view images to a single cost volume. The Learnt Stereo Machine (LSM)~\cite{RN08} ensures end-to-end MVS reconstruction by differentiable projection and unprojection operations. The features are unprojected into 3D feature grids with known camera parameters, and 3D CNN then is used to detect the surface of the 3D object in the voxel. Both SurfaceNet and LSM utilize volumetric representation; nevertheless, they only reconstruct low-resolution objects and have a huge GPU memory consumption of 3D voxel; for example, they created the world grid at a resolution of $32\times32\times32$.\\
3D cost volume has its advantage in encoding camera parameters and image features. DeepMVS~\cite{RN07} generates a plane-sweep volume for each reference image, and an encoder-decoder structure with skip connections is used to aggregate the cost and estimate depths with fully-connected conditional random field (Dense-CRF)~\cite{RN27}.~\cite{RN09} built a 3D cost volume by differentiable homography warping. Its memory requirement grows cubically with the depth quantization number, which makes it unrealistic for large scale scenes. The state-of-the-art method, R-MVSNet~\cite{RN10}, regularized 2D cost maps sequentially across depths via a convolutional gated recurrent unit (GRU)~\cite{RN28} instead of 3D CNNs, which reduced the memory consumption and made high-resolution reconstruction possible. However, R-MVSNet regularized the cost maps with a small $3\times3$ receptive field in the GRUs and down-sampled the output depth four times, which resulted in contextual information loss and coarse reconstruction.\\
Our RED-Net approach follows the idea of sequentially processing 2D features along the depth direction for wide-depth range inference. However, we introduce a recurrent encoder-decoder architecture to regularize the 2D cost maps rather than simply stacking the GRU blocks as in~\cite{RN10}. The RED structure provides multi-scale receptive fields to exploit neighborhood information effectively in fine resolution scenes, which allows us to achieve large-scale and full-resolution reconstruction with higher accuracy and efficiency and lower memory requirements.
\section{WHU Dataset}
This section describes the synthetic aerial dataset we created for large-scale and high-resolution Earth surface reconstruction call the WHU dataset. The aerial images in the dataset were simulated from a 3D surface model that was produced by software and refined by manual editing. The dataset includes a complete aerial image set and cropped sub-image sets for facilitating deep learning.
\subsection{Data Source}
A 3D digital surface model (DSM) with OSGB format~\cite{RN29} was reconstructed using Smart3D software~\cite{RN02} from a set of multi-view aerial images captured from an oblique five-view camera rig mounted on an unmanned aerial vehicle (UAV). One camera was pointed straight down and the optical axis of the other four surrounding cameras was at a 40$^{\circ}$ tilt angle, which guaranteed most of the scenes, including the building fa\c cade, could be well captured. We manually edited some errors in the surface model to improve its resemblance to the real scene. The model covered an area of about $6.7\times2.2$ km$^2$ over Meitan County, Guizhou Province in China with about 0.1 m ground resolution. The county contains dense and tall buildings, sparse factories, mountains covered with forests, and some bare ground and rivers.
\subsection{Synthetic Aerial Dataset}
First, a discrete 3D points set on a $0.06\times0.06\times0.06$ m$^3$ grid covering the whole scene was generated by interpolating the OSGB mesh. Each point includes the object position \emph{(X, Y, Z)} and the texture \emph{(R, G, B)}.\\
Then, we simulated the imaging process of a single-lens camera. Given the camera's intrinsic parameters (focal length \emph{f}, principal point \emph{x$_0$}, \emph{y$_0$}, image size \emph{W}, \emph{H}, and sensor size) and the exterior orientation (camera center \emph{(Xs, Ys, Zs)} and three rotational angles ($\varphi$, $\omega$, $\kappa$)). We projected the 3D discrete points onto the camera to obtain a virtual image, and the depth map was simultaneously retrieved from the 3D points. Note that the depth map was complete even on the building fa\c cade since the 3D model had full scene mesh.\\
The virtual image was taken at 550 m above the ground with 10 cm ground resolution. A total of 1,776 images ($5376\times5376$ in size) were captured in 11 strips with 90\% heading overlap and 80\% side overlap, with corresponding 1,776 depth maps as ground truth. We set the rotational angles at (0,0,0), and two adjacent images therefore could be regarded as a pair of epipolar images. A total of 1,760 disparity maps along the flight direction also were provided for evaluating the chosen stereo matching methods. We provided 8-bit RGB images and 16-bit depth maps with the lossless PNG format and text files that recorded the orientation parameters that included the camera center \emph{(Xs, Ys, Zs)} and the rotational matrix {\bf R}.
\subsection{Sub-Dataset for Deep Learning}
\begin{figure}
\begin{center}
\includegraphics[width=0.9\linewidth]{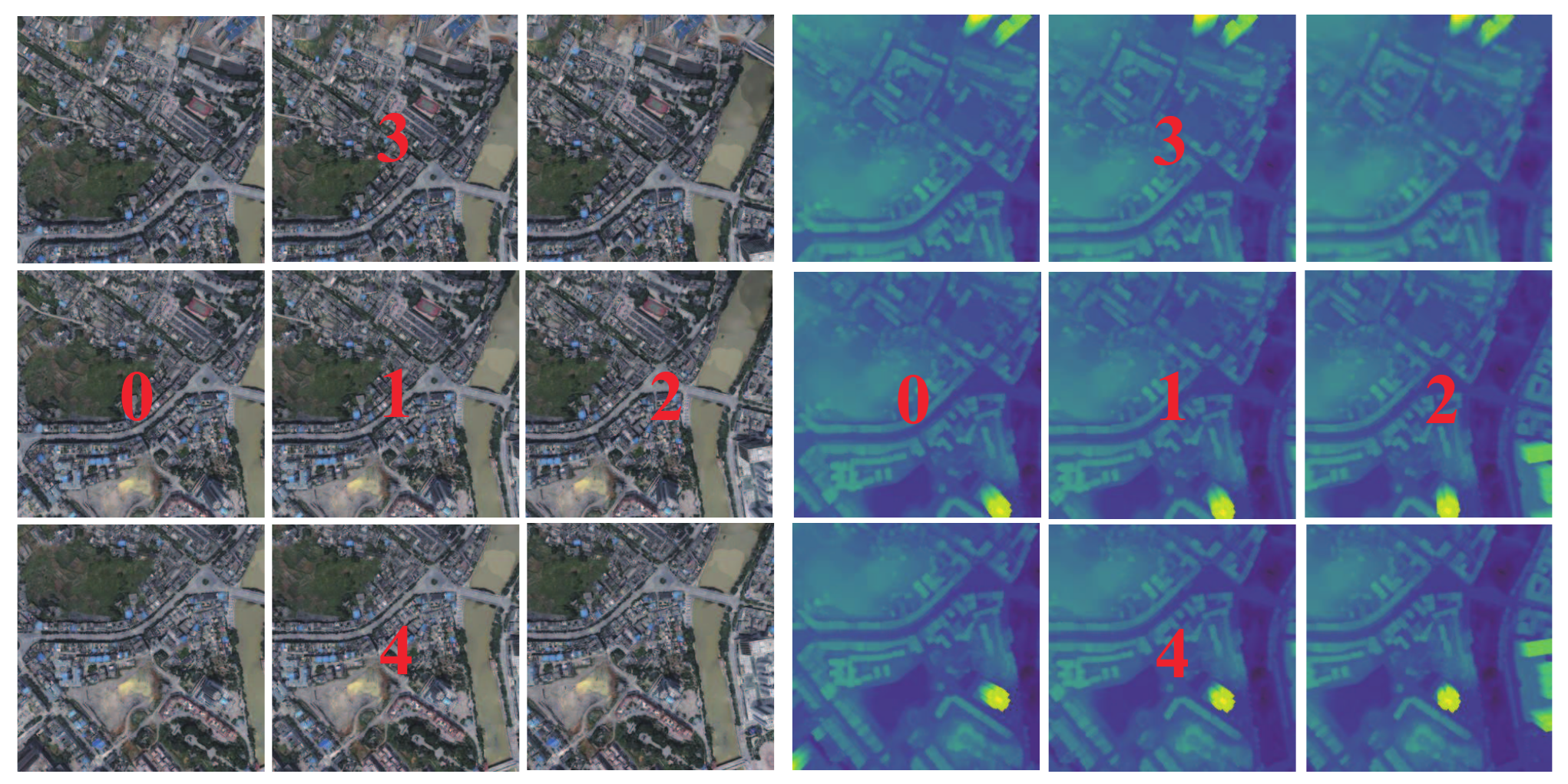}
\end{center}
\setlength{\abovecaptionskip}{-0.3cm}
   \caption{The images and depth maps from different viewpoints. A five-view unit took the Image with ID 1 as the reference image, the images with ID 0 and 2 in the heading direction and the images with ID 3 and 4 in the side strips as the search images. The three-view set consisted of images with ID 0, 1, and 2. In the stereo dataset, Image 1 and Image 2 were treated as a pair of stereo epipolar images.}
\label{fig:2}
\vspace{-0.3cm}
\end{figure}
\begin{figure}[ht]
\begin{center}
\subfloat[]{
\includegraphics[width=0.7\linewidth]{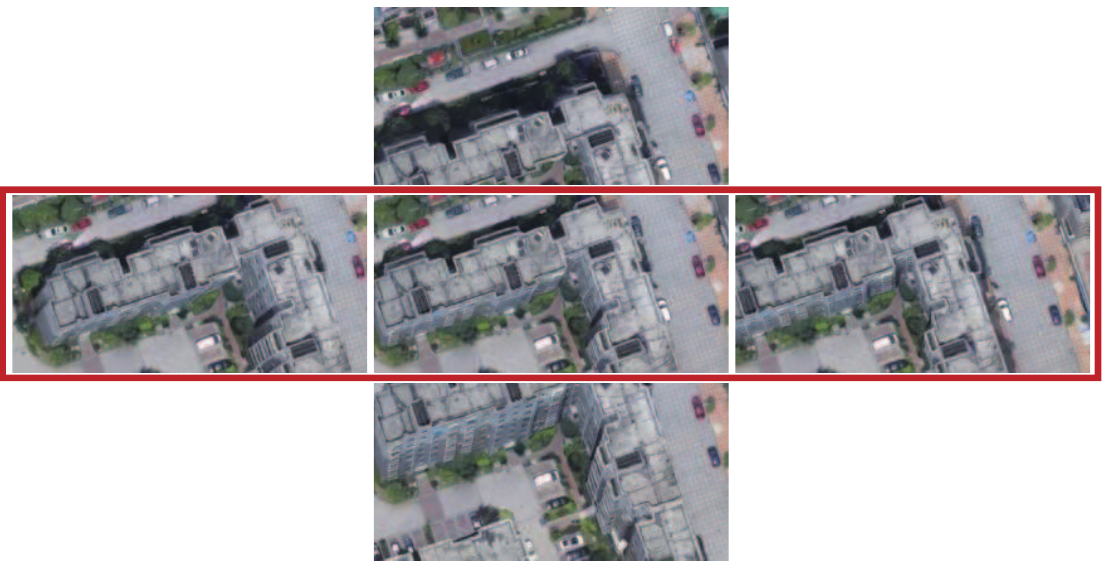}}
\setlength{\abovecaptionskip}{-0.4cm}
\label{3(a)}
\vspace{-0.4cm}
\subfloat[]{
\includegraphics[width=0.55\linewidth]{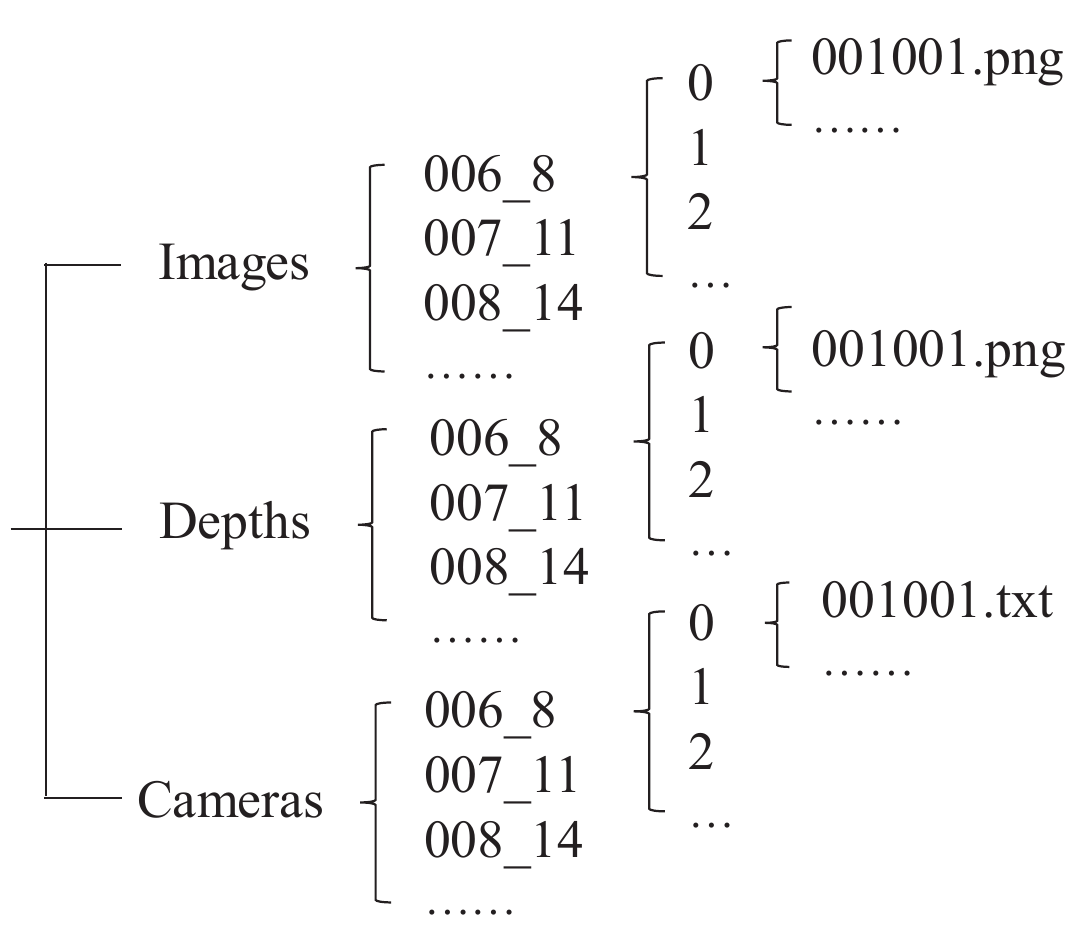}}
\setlength{\abovecaptionskip}{-0.4cm}
\label{3(b)}
\end{center}
\setlength{\abovecaptionskip}{-0.3cm}
   \caption{(a) A five-view sub-set with size of $768\times384$ pixels. The three sub-images in red rectangle comprise the three-view set. (b) The organization of images, depths, and camera files in the MVS dataset. }
\label{fig:3}
\vspace{-0.5cm}
\end{figure}
In addition to providing the complete dataset, we selected six representative sub-areas covering different scene types as training and test sets for deep learning methods, which are shown in Figure \ref{fig:1}. ``Area 1'' is a flat suburb with large and low factory buildings. ``Area 2'' contains trees, roads, buildings, and open spaces. ``Area 3'' is a residential area with a mixture of low and high buildings. ``Area 4'' and ``Area 5'' are the town center covering dense buildings with complex rooftop structures. ``Area 6'' is a mountainous area covered by agricultural land and forests. A total of 261 virtual images of Areas 1/4/5/6 were used as the training set, and 93 images from Area 2 and Area 3 comprised the test set. The ratio of the training to the test set was roughly 3:1. For a direct application of the deep learning-based MVS methods on the sub-dataset, we additionally provided a multi-view and a stereo sub-set by cropping the virtual aerial images into sub-blocks as an image of $5376\times5376$ pixels may not be fed into a current single GPU.\\
{\bf Multi-view Dataset}. A multi-view unit consists of five images as shown in Figure \ref{fig:2}. The central image with ID 1 was treated as the reference image, and the images with ID 0 and 2 in the heading direction and the images with ID 3 and 4 in the side strips were the search images. We cropped the overlapped pixels into the sub-block at a size of $768\times384$ pixels. A five-view unit yielded 80 pairs (400 sub images) (Figure \ref{fig:3}(a)). The depth maps were cropped at the same time. The dataset was ultimately organized as Figure \ref{fig:3}(b). The virtual images, depth maps, and camera parameters were in the first level folder. The second level folders took the name of the reference image in a five-view unit; for example, 006\_8 represented the eighth image in the sixth strip. The five sub-folders were named as 0/1/2/3/4 to store the sub images generated from the five-view virtual images respectively. In addition, there was a three-view dataset that consisted of the images with ID 0, 1, and 2.\\
{\bf Stereo Dataset}. Each adjacent image pair in a strip was also epipolar images. Similar to the multi-view set, we cropped each image and disparity map into $768\times384$ pixels and obtained 154 sub-image pairs in a two-view unit.
\section{RED-Net}
\begin{figure*}
\begin{center}
\includegraphics[width=0.9\linewidth]{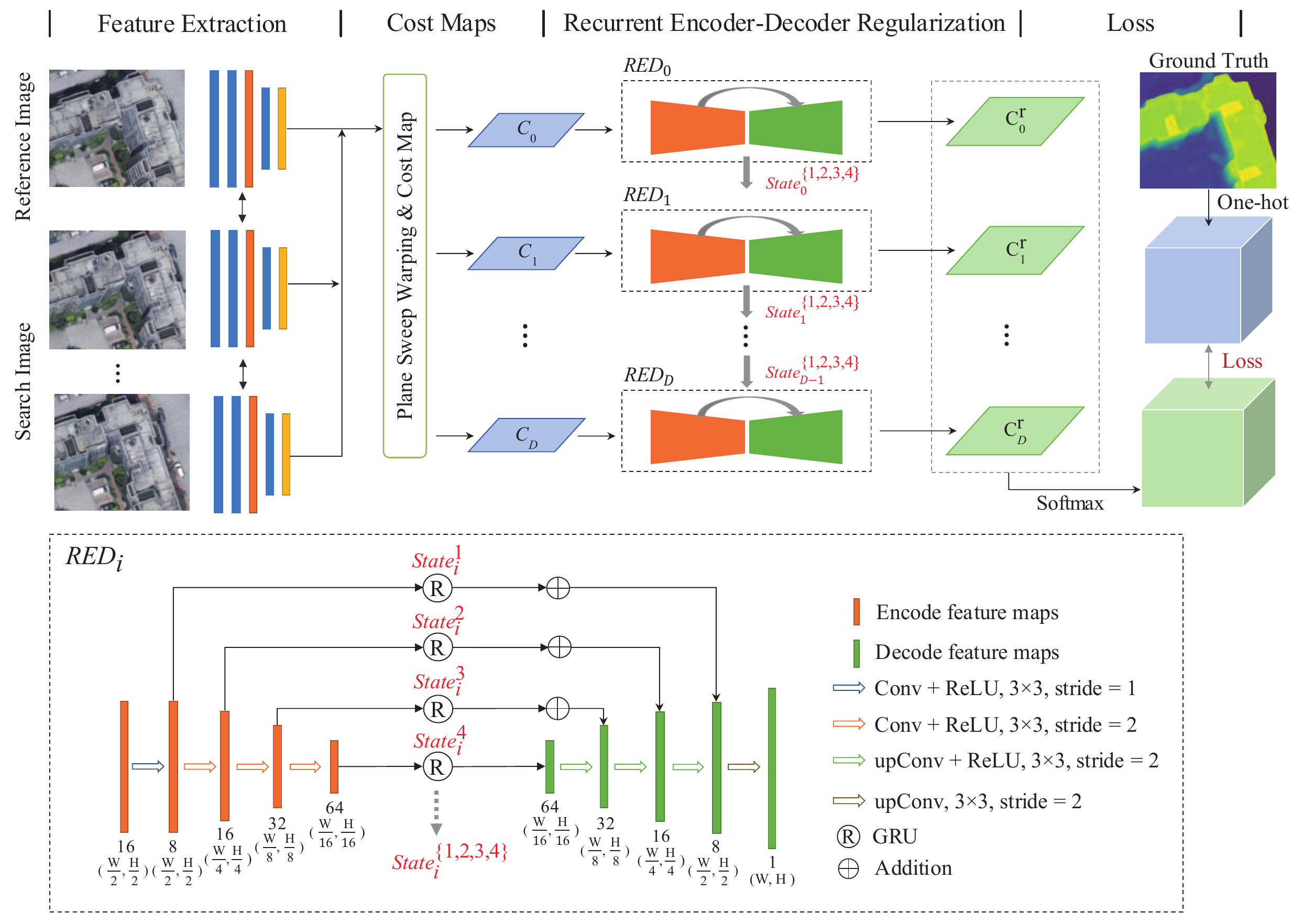}
\end{center}
\setlength{\abovecaptionskip}{-0.3cm}
   \caption{The structure of the RED-Net. \emph{W}, \emph{H}, and \emph{D} are the image width, height, and depth sample number, respectively.}
\label{fig:4}
\vspace{-0.3cm}
\end{figure*}

We developed a network, which we named RED-Net, that combines a series of weight-shared convolutional layers that extract the features from separate multi-view images and recurrent encoder-decoder (RED) structures that sequentially learn regularized depth maps across both the depth and spatial directions for large-scale and high-resolution multi-view reconstruction. The framework was inspired by~\cite{RN10}. However, instead of using a stack of three GRU blocks, we utilized a 2D recurrent encoder-decoder structure to sequentially regularize the cost maps, which not only significantly reduced the memory consumption and greatly improved the computational efficiency, but also captured the finer structures for depth inference. The output of RED-Net has the same resolution as the input reference images rather than being downsized by four as in~\cite{RN10}, which ensures high-resolution reconstruction for large-scale and wide depth range scenes. The network structure is illustrated in Figure \ref{fig:4}.\\
{\bf 2D Feature Extraction}. RED-Net infers a depth map with depth sample number \emph{D} from \emph{N}-view images where \emph{N} is typically no less than three. The 2D convolution layers first are separately used to extract the features of the \emph{N} input images with shared weights, which can be seen as an \emph{N}-way Siamese network architecture~\cite{RN30}. Each branch consists of five convolutional layers with 8, 8, 16, 16, 16 channels, respectively, and a $3\times3$ kernel size and a stride of 1 (except for the third layer, which has a $5\times5$ kernel size and a stride of 2). All of the layers are followed by a rectified linear unit (ReLU)~\cite{RN31} except for the last layer. The 2D network yields 16-channel feature representations for each input image half the width and height of the input image.\\
{\bf Cost Maps}. A group of 2D image features are back-projected onto successive virtual planes in 3D space to build cost maps. The plane sweep methods ~\cite{RN32} were adopted to warp these features into reference camera viewpoint, which is described as differentiable homography warping in~\cite{RN09,RN10}. The variance operation~\cite{RN09} was adopted to concatenate multiple feature maps to one cost map at a certain depth plane in 3D space. Finally, \emph{D} cost maps are built at each depth plane.\\
{\bf Recurrent Encoder-Decoder Regularization}. Inspired by the U-Net~\cite{RN33}, GRU~\cite{RN28}, and RCNN~\cite{RN34}, in this paper we introduce a recurrent encoder-decoder architecture to regularize the \emph{D} cost maps that are obtained from the 2D convolutions and plane sweep methods. In the spatial dimension, one cost map C$_i$ is the input to the recurrent encoder-decoder structure at a time, which is then processed by a four-scale convolutional encoder. Except for the first convolution layer with stride 1 and channel number 8, we doubled the feature channels at each downsampling step in the encoder. The decoder consists of three up-convolutional layers, and each layer expands the feature map generated by the previous layer and halves the feature channels. At each scale, the encoded feature maps are regularized by a convolutional GRU~\cite{RN10}, which are then added to the corresponding feature maps at the same scale in the decoder. After the decoder, an up-convolutional layer is used to upsample the regularized cost maps to the input image size and reduce channel number to 1.\\
In the depth direction, the contextual information of the sequential cost maps is recorded in the previous regulated GRUs and transferred to current cost map C$_i$. There are four GRU state transitions in the laddered encoder-decoder structure, denoted as \emph{state}, to gather and refine the contextual features in different spatial scales.\\
By regularizing the cost maps in the spatial direction and aggregating the geometric and contextual information in the depth direction by the recurrent encoder-decoder, RED-Net realized globally consistent spatial/contextual representations for multi-view depth inference. Compared to a stack of GRUs~\cite{RN10}, our multi-scale recurrent encoder-decoder exploits multi-scale neighborhood information with more details and less parameters.\\
{\bf Loss computation}. A cost volume is obtained by stacking all the regularized cost maps together. We turned it into a probability volume by utilizing a \emph{softmax} operator along the depth direction as accomplished in previous works~\cite{RN23}. From this probability volume, the depth value can be estimated pixel-wise and compared to the ground truth with the cross-entropy loss, which is the same as~\cite{RN10}.\\
To maintain an end-to-end manner, we did not provide a post-processing process. The inferred depth maps are translated into dense 3D points according to the camera parameters, all of which constitute the complete 3D scene. However, many classic post-processing methods~\cite{RN35} can be applied for refinement.
\begin{figure*}[htp]
\begin{center}
\includegraphics[width=1.0\linewidth]{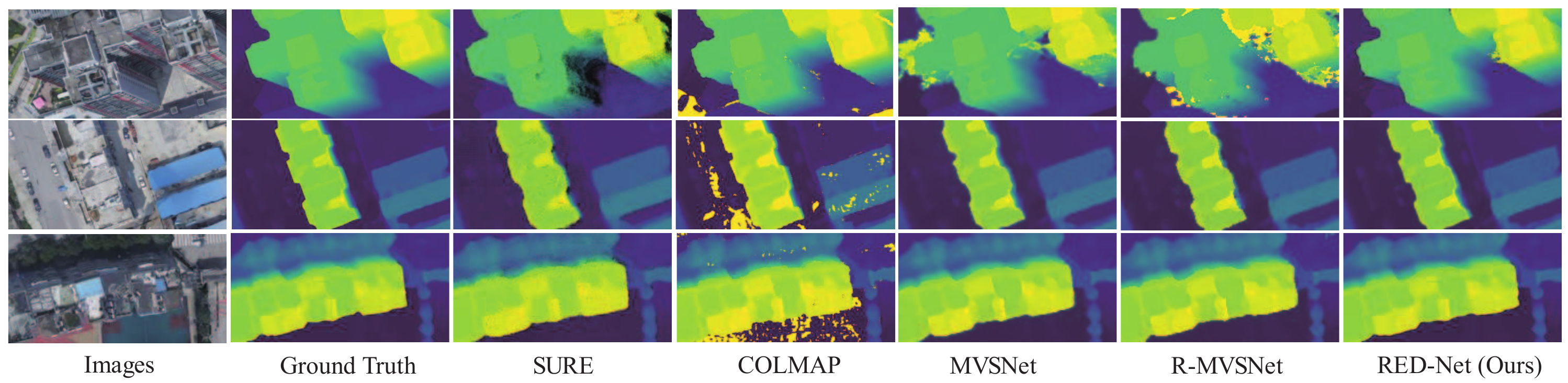}
\end{center}
\setlength{\abovecaptionskip}{-0.2cm}
   \caption{The inferred depth maps of three sub-units in the WHU test set. Our method produced the finest depth maps.}
\label{fig:5}
\vspace{-0.1cm}
\end{figure*}
\section{Experiments}
\subsection{Experimental Settings and Results}
We evaluated our proposed RED-Net on our WHU dataset and compared it to several recent MVS methods and software, including COLMAP~\cite{RN36} and commercial software SURE~\cite{RN03} (aerial version for trial~\cite{RN37}), which are based on conventional methods, and the MVSNet~\cite{RN09} and R-MVSNet~\cite{RN10}, which are based on deep neural networks. We directly applied COLMAP and SURE to the WHU test set, which contained 93 images ($5376\times5376$ in size) and output depth maps or dense clouds. We trained the CNN-based methods, which includes our method, with the WHU training set, which contained 3,600 sub-units ($768\times384$ in size) and then evaluated them on the WHU test set, which contained 1,360 sub-units with the same image size. The input view numbers were N=3 and N=5 for WHU-3 and WHU-5, respectively, with depth sample number D=200. The depth range can vary in each image, so we evaluated the initial depth with COLMAP and set the depth range accordingly for each image. In the test set, the depth number was variable and we set the interval at 0.15 m. The performances of the different methods were compared on the depth maps without any post-processing. For SURE, the generated dense point clouds were translated to depth maps in advance.\\
In the training stages of RED-Net, RMSProp~\cite{RN38} was chosen as the optimizer, and the learning rate was set at 0.001 with a decay of 0.9 for every 5k iterations. The model was trained for three epochs with a batch size of one, which involved about 150k iterations in total. All the experiments were conducted on a 24 GB NVIDIA TITAN RTX graphics card and TensorFlow platform.\\
We used four measures to evaluate the depth quality: 1) {\bf Mean absolute error (MAE)}: the average of the L1 distances between the estimated and true depths, and only the distances within 100 depth intervals were counted in order to exclude the extreme outliers; 2) {\bf $\textless$ 0.6m}: the percentage of pixels whose L1 error were less than the 0.6 m threshold; 3) {\bf 3-interval-error ($\textless$ 3-interval)}: the percentage of pixels whose L1 error was less than three depth intervals; 4) {\bf Completeness}: the percentage of pixels with the estimated depth values in the depth map.\\
Our quantitative results are shown in Table \ref{tab:1}. RED-Net outperformed all the other methods for all the indicators and obtained at least 50\% MAE improvement compared to the second-best R-MVSNet. For the 3-interval-error and 0.6 m threshold indicators, our method exceeded all the other methods at least 2\%. Our qualitative results in Figure \ref{fig:5} show that RED-Net's reconstructed depth map was the cleanest and most similar to the ground truth.
\begin{table}
\footnotesize
\begin{center}
\setlength{\tabcolsep}{2.0 mm}{
\begin{tabular}{c|ccccc}
\hline
\textbf{Method}   & \textbf{\begin{tabular}[c]{@{}c@{}}Train \\ \& Test\end{tabular}} & \textbf{\begin{tabular}[c]{@{}c@{}}MAE\\ (m)\end{tabular}} & \textbf{\begin{tabular}[c]{@{}c@{}}\textless{}3-interval\\ (\%)\end{tabular}} & \textbf{\begin{tabular}[c]{@{}c@{}}\textless{}0.6m\\ (\%)\end{tabular}} & \textbf{Comp.} \\ \hline
\textbf{COLMAP}   & / & 0.1548                                                     & 94.95                                                                         & 95.67                                                                   & 98\%           \\
\textbf{SURE}     & /                                                                 & 0.2245                                                     & 92.09                                                                         & 93.69                                                                   & 94\%           \\ \hline
\textbf{MVSNet}   & WHU-3                                                             & 0.1974                                                     & 93.22                                                                         & 94.74                                                                   & 100\%          \\
                  & WHU-5                                                             & 0.1543                                                     & 95.36                                                                         & 95.82                                                                   & 100\%          \\ \hline
\textbf{R-MVSNet} & WHU-3                                                             & 0.1882                                                     & 94.00                                                                         & 94.90                                                                   & 100\%          \\
                  & WHU-5                                                             & 0.1505                                                     & 95.64                                                                         & 95.99                                                                   & 100\%          \\ \hline
\textbf{RED-Net}  & WHU-3                                                            & 0.1120                                                     & 97.90                                                                         & \textbf{98.10}                                                                   & 100\%          \\
                  & WHU-5                                                             & \textbf{0.1041}                                            & \textbf{97.93}                                                                & 98.08                                                                   & 100\%          \\ \hline
\end{tabular}}
\end{center}
\setlength{\abovecaptionskip}{-0.1cm}
\caption{The quantitative results on WHU dataset.}
\label{tab:1}
\vspace{-0.3cm}
\end{table}
\subsection{GPU Memory and Runtime}
The GPU memory requirement and running speed of RED-Net, MVSNet, and R-MVSNet on the WHU dataset are listed in Table \ref{tab:2}. The memory requirement of MVSNet increased with depth sample number \emph{D}, whereas that of RED-Net and R-MVSNet were constant at \emph{D}. The occupied memory of RED-Net was nearly half that of R-MVSNet, and RED-Net could reconstruct a depth map with full resolution, which was 16-time larger than the latter.\\
The runtime was related to the depth sample number, input image size, and image number. Given the same \emph{N}-view images, (R-)MVSNet generated a depth map down-sampled by 4 and was slightly faster, while RED-Net kept the same resolution with input inference. Therefore, considering the output resolution, our network was much more efficient than the others.
\begin{figure*}
\begin{center}
\includegraphics[width=1.0\linewidth]{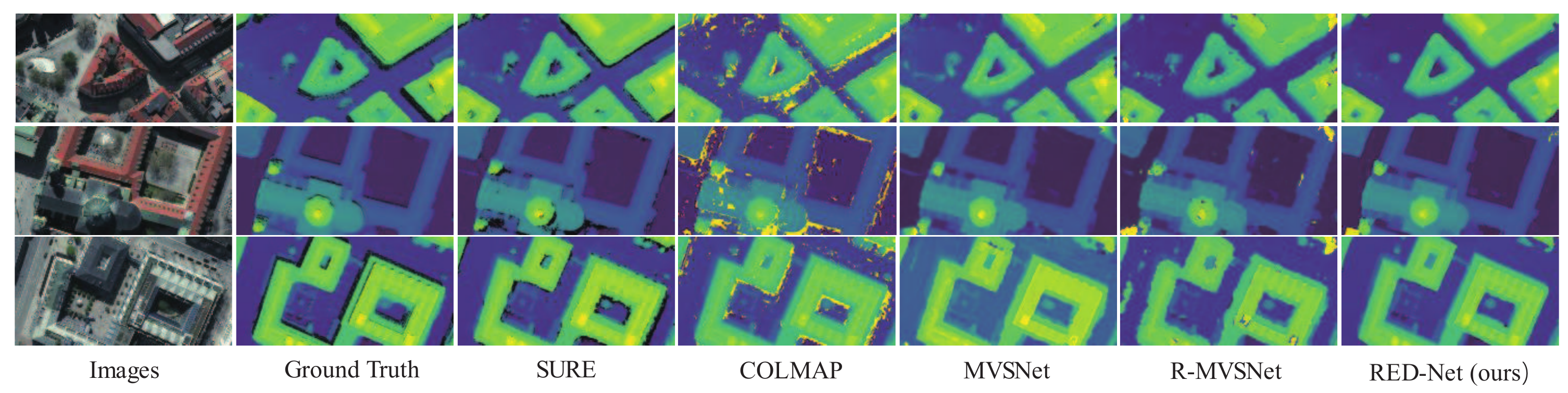}
\end{center}
\setlength{\abovecaptionskip}{-0.4cm}
   \caption{The inferred depth maps of three sub-units on M\"unchen aerial image set. The deep learning based methods are trained on the WHU-3 training set.}
\label{fig:6}
\vspace{-0.3cm}
\end{figure*}
\begin{table*}[]
\footnotesize
\begin{center}
\setlength{\tabcolsep}{3 mm}{
\begin{tabular}{c|c|cccc|c|c}
\hline
\multirow{2}{*}{\textbf{Methods}} &
  \multirow{2}{*}{\textbf{Input size}} &
  \multicolumn{4}{c}{\textbf{Depth sample number (3-view)}} &
  \textbf{(5-view)} &
  \multirow{2}{*}{\textbf{Output size}} \\ \cline{3-7}
         &         & D = 800     & D = 400     & D = 200    & D = 128    & D = 200    &         \\ \hline
\textbf{MVSNet}   & $384\times768$ & 17085M 1.1s & 8893M 0.6s  & 4797M 0.3s & 2749M 0.2s & 4797M 0.5s & $96\times192$  \\
\textbf{R-MVSNet} & $384\times768$ & 4419M 1.2s  & 4419M 0.6s  & 4419M 0.4s & 4419M 0.3s & 4547M 0.6s & $96\times192$  \\
\textbf{RED-Net}  & $384\times768$ & 2493M 1.8s  & 2493M 0.95s & 2493M 0.6s & 2493M 0.5s & 2509M 0.8s & $384\times768$ \\ \hline
\end{tabular}}
\end{center}
\setlength{\abovecaptionskip}{-0.2cm}
\caption{ Comparisons of memory requirement and runtime between (R-)MVSNet and RED-Net. Our method requires less memory but achieves full-resolution reconstruction.}
\label{tab:2}
\vspace{-0.3cm}
\end{table*}
\subsection{Generalization}
The WHU dataset was created under well-controlled imaging processes. To demonstrate the representation of the WHU dataset for aerial datasets and the generalization of RED-Net, five methods were tested on the real aerial dataset M\"unchen ~\cite{RN21}. The M\"unchen dataset is somewhat different from the WHU dataset in that it was captured at a metropolis instead of a town. It is comprised of 15 aerial images ($7072\times7776$ in size) and 80\% and 60\% overlapping in the heading and side directions, respectively. The three CNN-based models were pre-trained on the DTU or WHU datasets without any fine-tuning. The input view number of the M\"unchen dataset was N=3 and the depth sample resolution was 0.1 m. The quantitative results are shown in Table \ref{fig:3}. Some qualitative results are shown in Figure \ref{fig:6}.\\
Three conclusions can be drawn from Table \ref{tab:3}. First, RED-Net, which was trained on the WHU-3 dataset, performed the best in all the indicators. RED-Net also exceeded the other methods by at least 6\% in 3-interval-error. The model trained on the WHU-5 dataset performed almost the same as RED-Net. Second, the WHU dataset guaranteed the generalizability while the indoor DTU dataset could not. When trained on the DTU dataset, all the CNN-based methods performed worse than the two conventional methods. For example, (R-)MVSNet was 30\% worse than the two conventional methods in 3-interval-error; however, when trained on the WHU dataset, their performances were comparable to the latter. Finally, the recurrent encoder-decoder structure in RED-Net led to better generalizability compared to the stack of GRUs in R-MVSNet and the 3D convolutions in MVSNet. When trained on the DTU dataset, our method experienced a 20\% improvement over (R-)MVSNet in 3-interval-error.
\begin{table}
\footnotesize
\begin{center}
\begin{tabular}{c|cccc}
\hline
\textbf{Methods} &
  \textbf{Train set} &
  \textbf{\begin{tabular}[c]{@{}c@{}}MAE\\ (m)\end{tabular}} &
  \textbf{\begin{tabular}[c]{@{}c@{}}\textless{}3-interval\\ (\%)\end{tabular}} &
  \textbf{\begin{tabular}[c]{@{}c@{}}\textless{}0.6m\\ (\%)\end{tabular}} \\ \hline
\textbf{COLMAP}   & /     & 0.5860          & 73.36          & 81.95          \\
\textbf{SURE}     & /     & 0.5138          & 73.71          & 85.70          \\ \hline
                  & DTU   & 1.1696          & 43.19          & 61.26          \\
\textbf{MVSNet}   & WHU-3 & 0.6169          & 69.33          & 81.36          \\
                  & WHU-5 & 0.5882          & 70.43          & 83.46          \\ \hline
                  & DTU   & 0.7809          & 43.22          & 70.26          \\
\textbf{R-MVSNet} & WHU-3 & 0.6228          & 74.33          & 83.35          \\
                  & WHU-5 & 0.6426          & 74.08          & 83.68          \\ \hline
                  & DTU   & 0.6867          & 63.04          & 78.89          \\
\textbf{RED-Net}  & WHU-3 & \textbf{0.5063} & \textbf{80.67} & \textbf{86.98} \\
                  & WHU-5 & 0.5283          & 80.40          & 86.69          \\ \hline
\end{tabular}
\end{center}
\setlength{\abovecaptionskip}{-0.2cm}
\caption{Quantitative evaluation on the M\"unchen aerial image set with different MVS methods. The deep learning based methods were trained on the WHU or the DTU training set.}
\label{tab:3}
\vspace{-0.3cm}
\end{table}
\section{Discussion}
\subsection{Advantage of the Recurrent Encoder-Decoder}
 In this section, we evaluate the effectiveness of the recurrent encoder-decoder in an MVS network. We down-sampled the feature maps by four times in the 2D extraction stage. By doing this, the cost maps in RED-Net were the same size as R-MVSNet. The final output was also changed to 1/16 size of the input to keep consistent with the R-MVSNet. The results are compared in Table \ref{tab:4}. On the three aerial datasets, RED-Net demonstrated obvious advantages for all measures, which indicates that the high performance of RED-Net is not only due to improvement of the output resolution, but also to the encoder-decoder structure, which learned spatial and contextual representations better than stacked GRUs.
 \begin{figure*}[t]
\begin{center}
\includegraphics[width=0.9\linewidth]{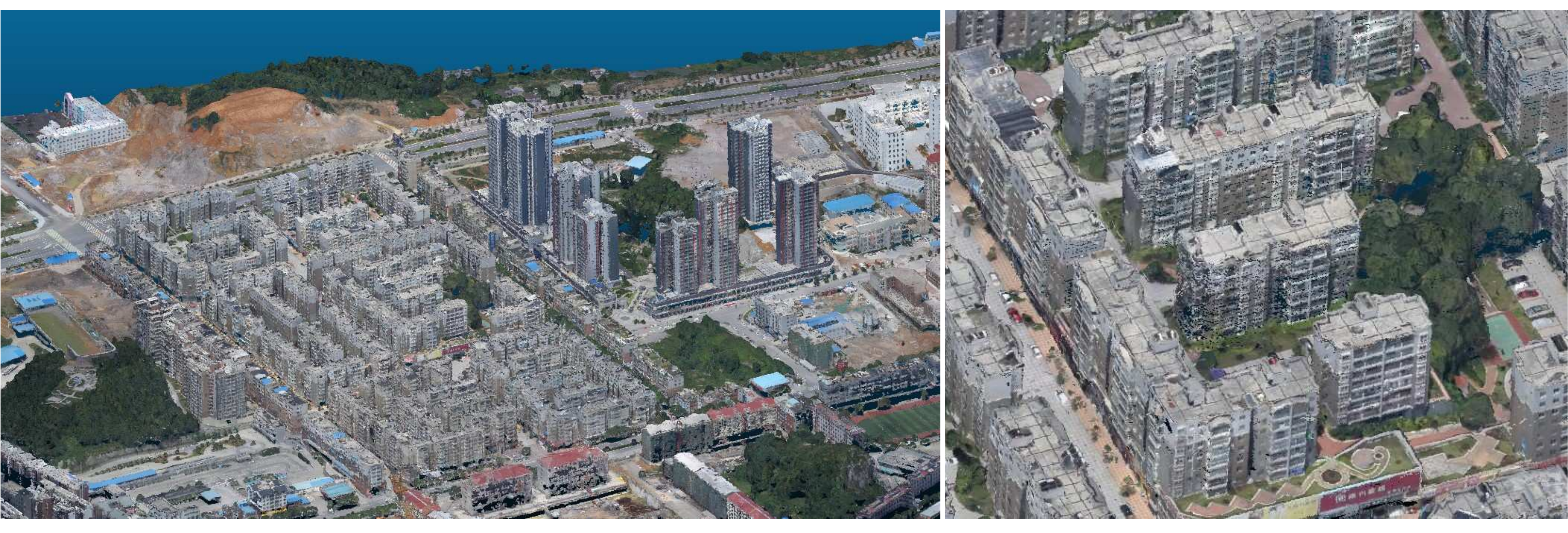}
\end{center}
\setlength{\abovecaptionskip}{-0.3cm}
   \caption{The point cloud reconstructions of a large area using RED-Net. The right is an enlarged part from the left scene.}
\label{fig:7}
\vspace{-0.2cm}
\end{figure*}
 \subsection{Evaluation on DTU}
 Although RED-Net is mainly developed for large-scale aerial MVS problem, it surpassed the state-of-the-art R-MVSNet on the close-range DTU dataset. Table \ref{tab:5} shows that, with the same post-processing (photometric and geometric filtering), the overall score of RED-Net \-outperformed that of R-MVSNet by 18\%, and also outperformed the results provided in~\cite{RN10} with full four post-processing methods. Overall score is derived from two representative indicators \emph{accuracy} and \emph{completeness} suggested by the DTU dataset~\cite{RN18} and used in~\cite{RN10}.
 \subsection{Large-scale Reconstruction}
 RED-Net produced full resolution depth maps with arbitrary depth sample numbers, which particularly can benefit high-resolution large-scale reconstruction of the Earth's surface from multi-view aerial images with a wide depth range. Moreover, RED-Net can handle three-view images with a size of $7040\times7040$ pixels on a 24GB GPU, taking only 58 seconds to infer a depth map with 128 depth sample numbers. When we inferred the depth of a scene covering $1.8\times0.85 $ km$^2$ (Figure \ref{fig:7}), RED-Net with 3-view input and 200 depth sample numbers took 9.3 minutes while SURE took 150 minutes and COLMAP took 608 minutes.
\section{Conclusion}
\noindent In this paper, we introduced and demonstrated a synthetic aerial dataset, called the WHU dataset, that we created for large-scale and high-resolution MVS reconstruction, which, to our knowledge, is the largest and only available multi-view aerial dataset. We confirmed in this paper that the WHU dataset will be a beneficial supplement to current close-range multi-view datasets and will help facilitate the study of large-scale reconstruction of the Earth's surface and cities.\\
We also introduced in this paper a new approach we \-developed for multi-view reconstruction called RED-Net. This new network was shown to achieve highly efficient large-scale and full resolution reconstruction with relatively low memory requirements, and its performance exceeded that of both the deep learning-based methods and commercial software. Our experiments also showed that RED-Net pre-trained on our newly created WHU dataset could be directly applicable to a somewhat different aerial dataset due to the proper training data and model's powerful generalizability, which has sent a signal that deep learning based approaches may take place of conventional MVS methods in practical large-scale reconstruction.
\begin{table}
\footnotesize
\begin{center}
\begin{tabular}{c|cccc}
\hline
\textbf{Dataset} &
  \textbf{Methods} &
  \textbf{\begin{tabular}[c]{@{}c@{}}MAE\\ (m)\end{tabular}} &
  \textbf{\begin{tabular}[c]{@{}c@{}}\textless{}3-interval\\ (\%)\end{tabular}} &
  \textbf{\begin{tabular}[c]{@{}c@{}}\textless{}0.6m\\ (\%)\end{tabular}} \\ \hline
\textbf{M\"unchen} & R-MVSNet & 0.4264 & 81.43 & 88.67 \\
\textbf{}        & RED-Net* & 0.3677 & 83.63 & 89.95 \\ \hline
\textbf{WHU-3}                & R-MVSNet & 0.1882 & 94.00 & 94.90 \\
                 & RED-Net* & 0.1574 & 95.52 & 96.03 \\ \hline
\textbf{WHU-5}            & R-MVSNet & 0.1505 & 95.64 & 95.99 \\
                 & RED-Net* & 0.1379 & 95.89 & 96.64 \\ \hline
\end{tabular}
\end{center}
\setlength{\abovecaptionskip}{-0.2cm}
\caption{Results of the R-MVSNet and RED-Net with the same size of inferred depth map on three datasets. `*' means that the cost maps and outputs of our method are downsampled by four as the R-MVSNet. Models are trained and tested on the same dataset respectively.}
\vspace{-0.2cm}
\label{tab:4}
\end{table}
\begin{table}
\footnotesize
\begin{center}
\setlength{\tabcolsep}{2.0 mm}{
\begin{tabular}{c|ccc}
\hline
\textbf{Methods(D=256)} & \textbf{Mean Acc.} & \textbf{Mean Comp.} & \textbf{Overall(mm)} \\ \hline
\textbf{R-MVSNet} {[}10{]} & 0.385 & 0.459 & 0.422 \\
\textbf{R-MVSNet}*         & 0.551 & 0.373 & 0.462 \\
\textbf{RED-Net}           & 0.456 & 0.326 & 0.391 \\ \hline
\end{tabular}}
\end{center}
\setlength{\abovecaptionskip}{-0.2cm}
\caption{Results of the R-MVSNet and RED-Net on DTU benchmark. `*' means our implementation with only photometric and geometric filtering post-processing, the same as in RED-Net.}
\label{tab:5}
\vspace{-0.3cm}
\end{table}
\section*{Acknowledgement}
This work was supported by the Huawei Company, Grant No. YBN2018095106.
{\small
\bibliographystyle{ieee_fullname}
\bibliography{egbib}
}
\end{document}